# Robot Design With Neural Networks, MILP Solvers and Active Learning

**Sanjai Narain, Emily Mak, Dana Chee, Todd Huster, Jeremy Cohen**
Perspecta Labs
snarain@perspectalabs.com

**Kishore Pochiraju, Brendan Englot**
Stevens Institue of Technology

**Niraj K. Jha**
Princeton University

**Karthik Narayan**
Starfruit LLC

**Abstract:** Central to the design of many robot systems and their controllers is solving a constrained blackbox optimization problem. This paper presents CNMA[1], a new method of solving this problem that is conservative in the number of potentially expensive blackbox function evaluations; allows specifying complex, even recursive constraints directly rather than as hard-to-design penalty or barrier functions; and is resilient to the non-termination of function evaluations. CNMA leverages the ability of neural networks to approximate any continuous function, their transformation into equivalent mixed integer linear programs (MILPs) and their optimization subject to constraints with industrial strength MILP solvers. A new learning-from-failure step guides the learning to be relevant to solving the constrained optimization problem. Thus, the amount of learning is orders of magnitude smaller than that needed to learn functions over their entire domains. CNMA is illustrated with the design of several robotic systems: wave-energy propelled boat, lunar lander, hexapod, cartpole, acrobot and parallel parking. These range from 6 real-valued dimensions to 36. We show that CNMA surpasses the Nelder-Mead, Gaussian and Random Search optimization methods against the metric of number of function evaluations.

[1] CNMA stands for Constrained optimization with Neural networks, MILP and Active learning

This research was developed with funding from the Defense Advanced Research Projects Agency (DARPA).

## Table of Contents

## 1. Introduction

Design tools for robots, such as simulators, are often only available as blackboxes with complex nonlinear relationships between inputs and outputs. The only service a blackbox provides is accepting an input and computing an output. The algorithms and mathematical models inside a blackbox are not available for analysis. Thus, central to the design of robots is solving a constrained blackbox optimization problem: find values of $x, y$ that optimize $\phi(x, y)$ such that $F(x) = y \wedge P(x, y)$ where $F$ is a potentially nonlinear function available as a blackbox, $x$, $y$ are vectors of discrete and continuous variables, $\phi$ is a potentially nonlinear function and $P$ is potentially nonlinear constraint. $P$ can be a conjunction of several constraints. This paper presents CNMAfor solving this problem in which $\phi, P$ are linear. Nonlinear $\phi, P$ are easily accommodated by moving nonlinearities into the definition of $F$ as shown in Section 8. CNMA is conservative in the number of potentially expensive function evaluations, allows specifying complex constraints directly rather than as hard-to-design penalty functions and is resilient to non-termination of function evaluations. Constraints can be recursive: they can simultaneously govern both input and output.

CNMA samples points in the domains of the function, and learns a neural network surrogate of the function. It transforms that surrogate into an equivalent Mixed Integer Linear Program [12] and constructs a conjunction of this with $P(x, y)$. It optimizes $\phi(x, y)$ subject to this conjunction using industrial-strength MILP (Mixed Integer Linear Programming) solvers e.g., [4]. It then checks the solution for correctness. If correct, CNMA outputs the solution. If incorrect, CNMA computes a new training instance from the solution and restarts. This "learning-from-failure" step has the effect of sampling the region of the domain relevant to solving the optimization problem. Thus, it reduces the number of function evaluations by orders of magnitude compared to that needed for learning the function over its entire domain.

CNMA is resilient to failure of function evaluation by leveraging the resilience of neural networks to missing information. If a function cannot be evaluated at a point, CNMA just samples another point. Constraints are represented naturally as constraints in an MILP language, not indirectly as hard-to-design penalty or barrier functions [6, 5] as with constrained optimization methods [6,7,8].

CNMA is illustrated by the design of diverse robotics systems and their controllers: wave-energy propelled boat, system-controller codesign for the OpenAI/Gym lunar lander, hexapod gait repair, cartpole, acrobot and a parallel parking maneuver. In addition, a benchmark nonlinear optimization problem is solved. These problems range in real-valued dimensions from 6 to 36 and are governed by complex constraints, even recursive ones that span both input and output variables. Against the metric of the number of function evaluations, CNMA is shown to surpass the following traditional optimization tools: Nelder-Mead, Random Search and Gaussian processes.

## 2. Relationship with previous work

For constrained blackbox optimization, derivative-based methods e.g. [6] tend to be infeasible when the cost of evaluating functions is high since derivative computation significantly increases the number of function evaluations. Derivative-free methods e.g. [8, 22, 18] are conservative in the number of times they evaluate functions. They build surrogate functions based on current samples and use those to compute the best point to sample next. However, their complexity can be high. Gaussian processes have time complexity $O(n^3)$ where $n$ is the sample size [3]. Some methods are not resilient to failure to evaluate the function for certain inputs, as can happen, for example, with computational fluid dynamics simulators [21]. An artificial value has to be assigned to the function resulting in a distorted surrogate function. Constraints are modeled as penalty or barrier functions encoding the cost of violating them. Then, unconstrained optimization is used to minimize this cost. However, construction of penalty functions requires substantial creativity on the part of the user [6, 7]. Constraint solving is treated as another optimization problem so the special nature of constraints is not leveraged for their solution. The dReal SMT solver [19] solves nonlinear constraints but not when function definitions are only available as blackboxes.

CNMA is simpler to use since constraints can be directly specified in a MILP language. There is no need to design penalty functions. The MILP solver also leverages the special nature of constraints to efficiently solve these. Like derivative-free methods, CNMA also uses a surrogate function to compute the next sample to evaluate in search of a solution. The difference is that the surrogate function is a neural network and its equivalent MILP is used to suggest the next relevant point to sample. Since MILP solvers are optimized to solving constraint on a large scale, they are a natural candidate for evaluating relevance to the *constraint* part of the problem. CNMA uses this idea to learn from failure. Thereby, the number of samples needed to solve problems is many orders of magnitude less than that needed to learn the functions over their entire domain. An extreme case is the 36-dimensional hexapod where CNMA starts with just two initial samples to compute a speed that is hard or impossible to surpass by traditional solvers. See Section 7.

At present, CNMA does not determine that a constraint is unsolvable. It keeps learning functions in new parts of their domains in the hope that more information would let it find a solution. Future research problems include finding an appropriate neural network architecture, detecting constraint unsolvability and its root cause; a theoretical understanding of CNMA; multiple-function CNMA; parallel CNMA; and neural network compression e.g., [17] to improve the efficiency of MILP solving.

## 3. Preliminaries

A mixed-integer linear constraint is of the form $a_0 * x_0 + \ldots + a_k * x_k \leq b$ where $a_i, b$ are real-valued constants and the $x_i$ are real-valued or integer-valued. A MILP is a set of such constraints with a linear objective function $\phi(v_0, .., v_m)$ where each $v_i$ is a variable appearing in a constraint. A MILP solver will find values of all variables in the program optimizing the function while satisfying all constraints. CNMA learns fully-connected neural networks with the ReLU activation function. A neural network has one input layer, one output layer and zero or more hidden layers. The output layer can have multiple neurons allowing modeling of multi-output functions. The ReLU activation function is simply $max(0, x)$ but it gives neural networks the power to learn nonlinear functions. However, $max(0, x)$ is can be modeled as a MILP using the big M method [12] . By also modeling the overall system requirement as another mixed integer linear constraint, scalable MILP solvers can be used to efficiently solve the neural network surrogate along with the requirement.

## 4. CNMA definition

CNMA solves the problem of finding $x, y$ that optimize $\phi(x, y)$ such that $F(x) = y \wedge P(x, y)$ where $F$ is a nonlinear function, $x$, $y$ are vectors of discrete and continuous variables, $\phi$ is a linear function and $P$ is a linear constraint. $F$ is called the forward function. Nonlinear $\phi$ and $P$ are easily accommodated by moving their nonlinearity into the definition of $F$. See Section 8. As shown in Figure 1, CNMA constructs a training set $T$ by evaluating $F$ for different samples in the domain of $F$ and labeling them with associated outputs. $T$ is used to learn an approximation $G$ to $F$ in the form of a neural network, with the ReLU activation function. The neural network is then converted into an equivalent MILP using the method of [12]. An MILP solver is used to find values of $x$ and $y$ that optimize $\phi(x, y)$ subject to the conjunction of the MILP and $P(x, y)$. Let these values be $x_0, y_0$ respectively. CNMA then checks whether $P(x_0, F(x_0))$ holds and whether $\phi(x_0, F(x_0))$ is acceptable e.g., above or below a desired threshold. If so, CNMA outputs as solution $x = x_0, y = F(x_0)$. Otherwise, CNMA adds $(x_0, F(x_0))$ to $T$ and retries. If the MILP solver fails to produce a solution, or $F(x_0)$ fails to terminate, CNMA selects a point $z$ in the domain of $F$, adds $(z, F(z))$ to the training set, and retries. If $F(z)$ does not terminate, a new $z$ is selected and this step repeated. The addition of $(x_0, F(x_0))$ is a form of learning from error that has the effect of learning that part of the domain of $F$ relevant to solving $P(x, y)$. Thus, the sampling of $F$ is reduced by many orders of magnitude over the fine-grained sampling needed to learn $F$ over its entire domain. Algorithm 1 precisely defines this plan.

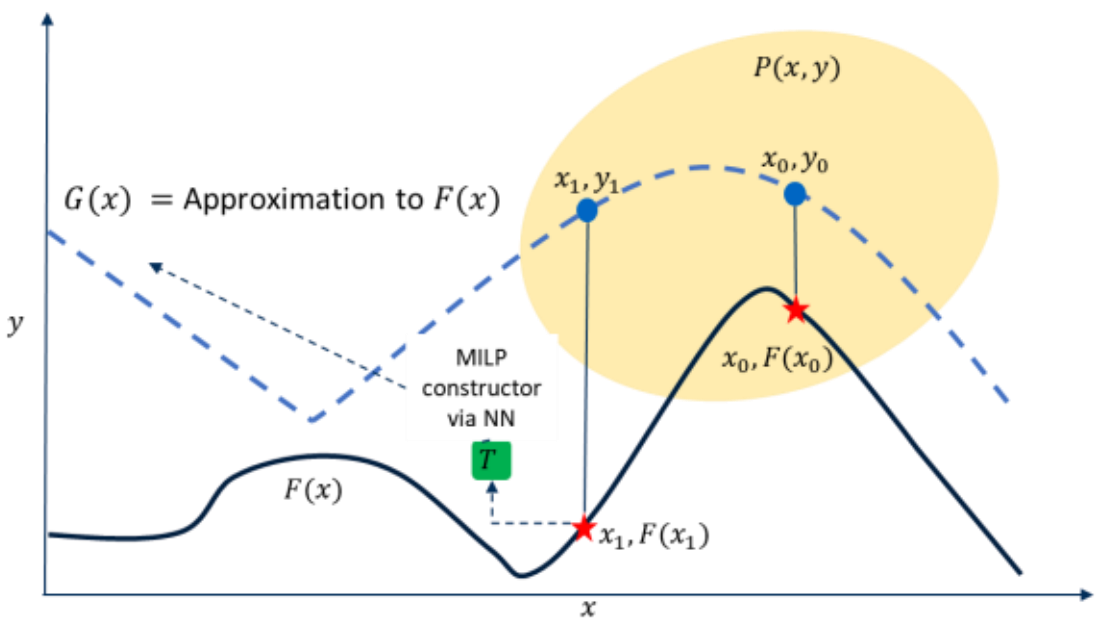

Figure 1. Single Function CNMA Overview

**Algorithm 1** CNMA, Single Forward Function

**Input:** a problem definition of the form $\max_{x,y} \phi(x,y)$ s.t. $y = F(x) \wedge P(x,y)$, a method to randomly sample candidate solutions $x \in X$, maximum number $N$ of CNMA iterations. $\phi, P$ are linear, $x, y$ are vectors of discrete and continuous variables, and $F$ is a potentially nonlinear function available as a blackbox.

**Output:** a solution, $(x^*, y^*)$, to the above problem

**function** CNMA($\phi, F, P, X, N$)
- $samples \leftarrow \{(x_i, F(x_i))\}_{i=1,2,\cdots,n}$ where $x_i$ denotes a random sample drawn from $X$
- $solutions \leftarrow$ empty list
- **for** $i = 1, 2, \cdots, N$:
  - $nn \leftarrow$ a fully connected ReLU regression network, i.e., with identity activation for the last layer, which takes in as input a vector $x \in X$ and attempts to predict $f(x)$; use $samples$ to train this neural network
  - $milp \leftarrow$ the mixed-integer linear program:
    $$\max_{x,y} \phi(x,y) \ \text{s.t.}\ nn_to_milp(nn) \wedge P(x,y)$$
  - $x^*, \hat{y} \leftarrow$ potentially infeasible solution to $milp$, obtained via a MILP solver
  - **if** $(x^*, \hat{y})$ is a feasible solution to $milp$:
    - $y^* \leftarrow F(x^*)$ // if $F$ does not terminate within some time limit then $F$ returns $\infty$
    - **if** $y^*$ is finite:
      - **if** $P(x^*, y^*)$ is satisfied:
        - Append $(x^*, y^*)$ to $samples$ // CNMA learns from success
        - Append $(x^*, y^*)$ to $solutions$
      - **else:**
        - Append $(x^*, y^*)$ to $samples$ // CNMA learns from failure
      - **endif**
    - **endif**
  - **else:** // the solution to $milp$ is infeasible
    - Append randomly drawn sample(s) $\{x_i, F(x_i)\}_{i=1,2,\cdots}$ to $samples$
  - **endif**
- **return** the best solution from $solutions$, sorting by $\phi(x,y)$

**endfunction**

## 5. Designing wave-energy-propelled marine robot

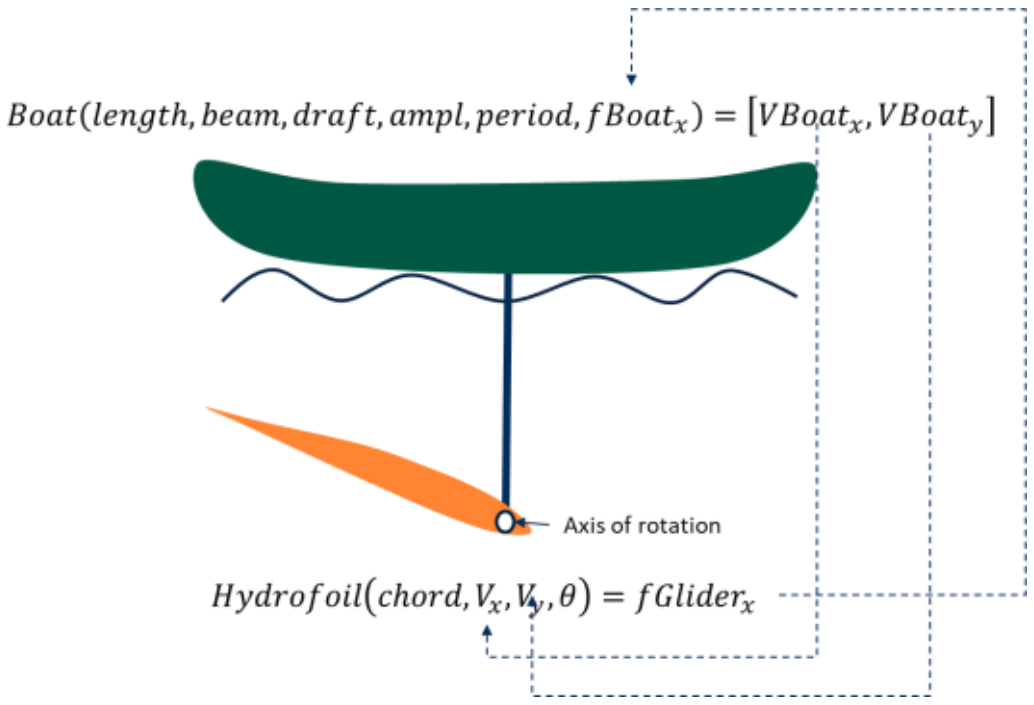

Figure 2. Boat Propelled By Wave Energy

As shown in Figure 2, the rise and fall of a boat floating on a wave pulls and pushes at a hydrofoil below, causing a rotation about an axis. This rotation generates forward force during both the upward and downward wave motion, much like when a swimmer flaps flippers in a pool, his body is propelled forward. The marine robot is inspired by the Wave Glider [13]. The design problem is computing the dimensions of the boat and hydrofoil to maximize the steady-state forward sailing speed for a given wave condition. The equilibrium constraints are that the force generated by the hydrofoil equals that applied to the boat and that the glider and boat velocities are equal. An additional constraint is that the magnitude of the horizontal velocity be higher than that of the vertical. Note the recursive relationship between the variables: force is an output of $Hydrofoil$ but an input to $Boat$ whereas velocities are outputs of the latter and inputs to the former.

The marine robot is modeled with two functions. The first is $Hydrofoil(ch, V_x, V_y, \theta) = fGlider_x$ that computes the forward force output by a hydrofoil of length $ch$, moving through water at velocity $V_x, V_y$ at an angle of attack $\theta$. This is the force it applies to the boat. It is implemented with the computational fluid dynamics package xfoil [21]. The second function is $Boat(length, beam, draft, ampl, period, fBoat_x) = [VBoat_x, VBoat_y]$ that outputs the steady-state forward speed of a boat given its 3D dimensions $length, beam, draft$, the amplitude $ampl$ and $period$ of the wave, and the forward force $fBoat_x$ applied to it by the hydrofoil. This function is computed by a program encoding a solution to a differential equation. The first two equilibrium constraints are enforced by eliminating $V_x$ and $V_y$ and consolidating the two functions into the CNMA forward function $F(x) = y$ where:

$$x = [length, beam, draft, ampl, period, fBoat_x, ch, \theta]$$
$$y = [VBoat_x, VBoat_y, fGlider_x]$$

$F$ calls $Boat$ to compute $VBoat_x$ and $VBoat_y$ and then inputs these to $Hydrofoil$ to compute $fGlider_x$. The third equilibrium constraint is now $fBoat_x = fGlider_x$. To tolerate small force differences the equality is modeled as $|fBoat_x - fGlider_x| \leq \epsilon * fBoat_x$ where $\epsilon$ is set to 5%. Note that a constraint with an absolute value can be modeled as a pair of linear constraints [2]. Finally, $P(x, y) = (|fBoat_x - fGlider_x| \leq \epsilon * fBoat_x \wedge VBoat_x \geq VBoat_y)$ and $\phi(x, y) = VBoat_x$. CNMA solves this problem to produce a maximum boat speed of 3.755 m/sec, where $ampl$ and $period$ were set to 4 and 7 respectively. The solution was:

$$x = [0.2, \quad 0.2, \quad 0.2, \quad 4.0, \quad 7.0, \quad 104.0139, \quad 1.5, \quad 30.947]$$
$$y = [3.755, \ 2.285, \ 105.482]$$

We now show how CNMA works but, for brevity, only for design of the *Hydrofoil*. Given fixed chord length and velocity we need to find an angle of attack that produces a forward force (divided by 500) in the range [0.25, 0.4]. Formally, we solve $Hydrofoil(chord, Vx, V_y, \theta) = f_x \wedge P(chord, V_x, V_y, \theta, f_x)$ where $P(chord, Vx, Vy, \theta, f_x) = (chord = 0.5 \wedge V_x = 1 \wedge V_y = 1 \wedge \theta \in [20, 40] \wedge f_x / 500 \in [0.25, 0.4])$.

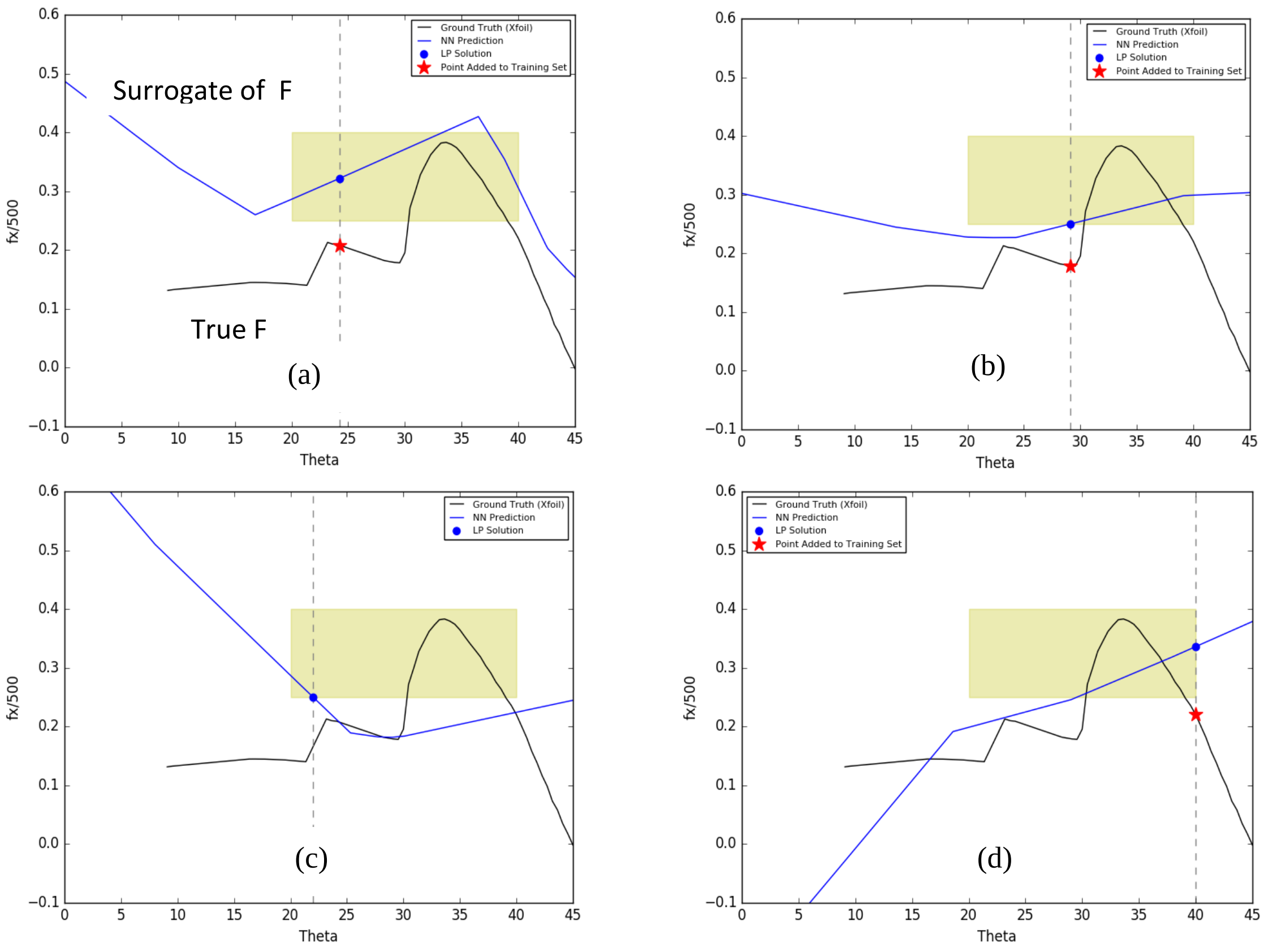

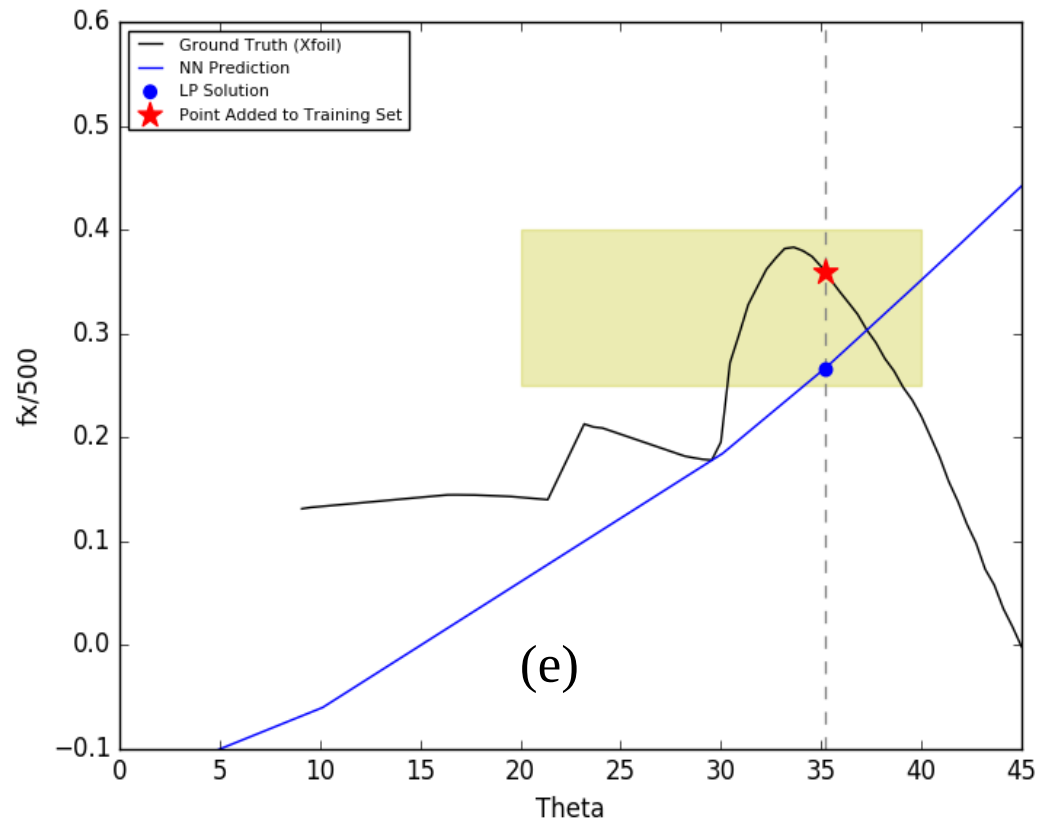

Figure 3. CNMA progress towards solution

Figures 3 (a)-(e) show how CNMA finds a solution in just 4 additional calls to $Hydrofoil$. The solid black curve is the plot of $\frac{f_x}{500}$ against $\theta$, computed by evaluating $Hydrofoil(chord = 0.5, V_x = 1, V_y = 1, \theta)$ for different values of $\theta$.The light green rectangle is the region where $P$ is true. A solution to the above constraint is any point on the black curve within the light green rectangle. CNMA first generates 22 samples in the domain of $Hydrofoil$ and evaluates $f_x$ for each to create a training set of instances such as:

| $chord$ | $V_x$ | $V_y$ | $\theta$ | $\frac{fx}{500}$ |
|---|---|---|---|---|
| 0.3375 | 2.5375 | 3.5125 | 39.375 | 1.747114 |
| 0.45625 | 1.31875 | 1.31875 | 30.9375 | 0.469039 |
| 0.753125 | 2.903125 | 2.415625 | 29.53125 | 5.601161 |
| 1.821875 | 1.684375 | 1.196875 | 26.71875 | 7.958876 |

CNMA then learns a neural network from this set to create a surrogate of $Hydrofoil$. The blue curve in Figure 3 (a) is the plot of $\frac{f_x}{500}$ against $\theta$ by evaluating this surrogate. This blue curve will change as new training instances are added by CNMA. CNMA converts this neural network into an MILP and solves it along with $P(chord, V_x, V_y, \theta, f_x)$ to produce the solution $chord = 0.5$, $V_x = 1$, $V_y = 1$, $\theta = 24.25, \frac{fx}{500} = 0.33$. In Figure 3 (a) this is the point on the blue curve in the light green rectangle. CNMA now checks this solution against the correct (not surrogate) definition of $Hydrofoil$ to produce $\frac{fx}{500} = 0.20$. As 0.20 is outside of [0.25, 0.4], it does not satisfy $P$. CNMA now adds the new training instance (0.5, 1, 1, 24.25, 0.20) to the training set and restarts. Figure 3 (b) shows the new surrogate function. While a solution is found, the true value of $\frac{fx}{500}$ for it still does not satisfy $P$. CNMA now adds the new training instance (0.5, 1, 1, 29.08, 0.17) and restarts. Figure 1 (c) shows the new surrogate function. While a solution is found, an attempt to find the true value of $\frac{fx}{500}$ does not terminate (because xfoil does not terminate). CNMA finds a random point in the domain of $Hydrofoil$, and computes the associated training instance (0.51, 1.74, 0.21, 6.72, 0.001) and restarts. Figure 3 (d) shows the new surrogate function. While a solution is found, the true value of $\frac{fx}{500}$ does not satisfy $P$. CNMA now adds one more training instance (0.5. 1, 1, 40, 0.22) and restarts. Figure 3 (e) shows the new surrogate function. This time a solution is found, and the true value of $\frac{fx}{500}$ does satisfy $P$. Thus, CNMA outputs the solution $chord = 0.5$, $Vx = 1$, $Vy = 1$, $\theta = 35.21$, $\frac{fx}{500} = 0.36$.

## 6. Robot system-controller codesign

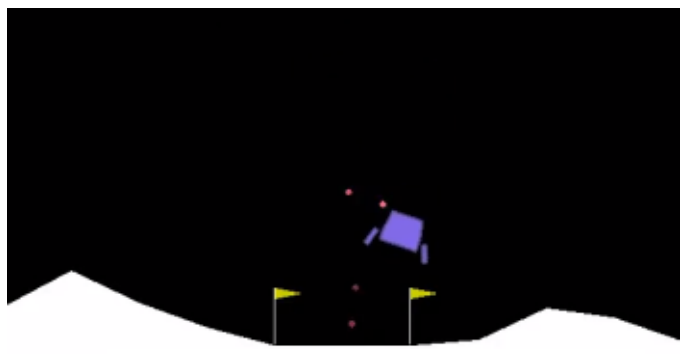

Figure 4. Lunar lander attempting to land on flat terrain between flagpoles

Robots and their controllers are often designed separately. This can lead to design inconsistency whose resolution can be time-consuming. If robots can be efficiently reconfigured and their controllers efficiently designed, we raise the possibility of system-controller codesign by solving a single optimization problem that encodes system and controller constraints and objectives. We could even customize the designs to specific operating conditions. We illustrate this concept by designing a lunar lander module and its controller by solving a single optimization problem in CNMA. The lander we use is based on that in OpenAI/Gym. We assume that the controller is PID-based, so the controller design problem reduces to finding optimal values for the PID coefficients. From a mother ship, the module is ejected with a certain force and then its engines fire both vertically and horizontally to guide it towards landing on the flat pad between two flagpoles. Our goal is to compute the system and controller design and initial position and force that would maximize the reward while satisfying constraints on a successful landing, time to land and fuel usage. It is not unrealistic to imagine that the lander system parameters, the PID parameters and the initial conditions are efficiently settable by mission control.

The input vector to the CNMA forward function $F$ is $x = [mep, sep, la, ld, lw, lh, lst, kp_alt,$

$kd_alt$, $kp_ang$, $kd_ang$, $initial_x, initial_y, initial_fx, initial_fy$]. The variables $mep, sep$, $la$, $ld$, $lw, lh$ $lst$ are system design parameters denoting, respectively, main engine power, side engine power, leg away length, leg down length, leg width, leg height, and leg spring torque. The variables $kp_alt, kd_alt$, $kp_ang$, $kd_ang$ are controller design parameters denoting, respectively, the P and D values for the vertical and horizontal engine controllers. The I coefficient is set to 0. The variables $initial_x, initial_y$, $initial_fx$, $initial_fy$ are initial condition parameters, denoting respectively, the initial position and force at the time of lunar lander ejection from the mother ship. The forward function $F(x) = y$ simulates the trajectory of the lander defined by its system design parameters, PID parameters and the initial conditions. It runs for a fixed number of time steps or until the lander lands or crashes. At each time step, it measures the "error" i.e., the distance between its current position and the landing point, and using the PID values computes engine firing actions to guide the lander towards the landing point. It outputs $y = [fuel, time, success, reward]$ denoting, respectively, the fuel used, time taken to land, whether the landing was safe, and the reward, a measure of the quality of the landing, as defined by OpenAI/Gym. The codesign problem is now to maximize the objective function $\phi(x) = reward$ subject to $F(x) = y \wedge P(x, y)$ where $P(x, y) = (success = 1 \wedge fuel <= 75 \wedge time \leq 10 \wedge Bounds(x, y))$. CNMA produced the following system design:

$$[mep,\ sep,\ la,\ ld,\ lw,\ lh,\ lst] =$$
$$[20.0, 5.0, 1.0, 33.43986077874244, 7.694512916653788, 5.540066165237896, 65.99264751099187]$$
$$\left[kp_{alt},\ kd_{alt},\ kp_{ant},\ kd_{ang}\right] =$$
$$[10.0, 6.237566661932409, 7.963989621392403, 3.982661097435201]$$
$$[initial_x, initial_y\ ] = [16.0, 14.869503942449763]$$
$$[initial_fx, initial_fy\ ] = [-2000.0, -1307.9525909825345]$$

With the associated output:

$$[success, fuel, time, reward] = [1.0, 67.76484, 4.38, 447.77823]$$

The bounds on the variables were as follows:

$$mep \in 1.0, 20.0]$$
$$sep \in [0.1, 5.0]$$
$$la \in [1.0, 40.0]$$
$$ld \in [1.0, 40.0]$$
$$lw \in [0.1, 10.0]$$
$$lh \in [0.1, 10.0]$$
$$lst \in [1.0, 100.0]$$
$$kp_alt \in [0.1, 10.0]$$
$$kd_alt \in [0.1, 10.0]$$
$$kp_ang \in [0.1, 10.0]$$
$$kd_ang \in [0.1, 10.0]$$
$$initial_x \in [14.0, 16.0]$$
$$initial_y \in [14.0, 15.0]$$
$$initial_fx \in [-2000.0, 2000.0]$$
$$initial_fy \in [-2000.0, 2000.0]$$

## 7. Designing hexapod gait

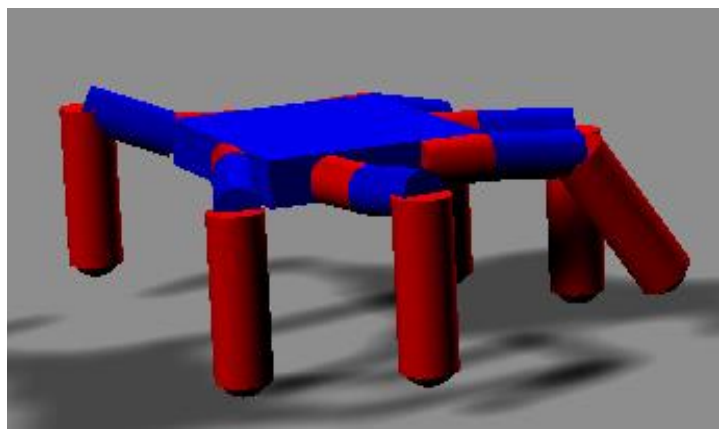

Figure 5. A hexapod

We now address a problem inspired by the work of [5] for adapting robot gait to failures in the field. The robot is 6-legged with each leg consisting of three segments. Associated with each leg $i$ is a vector of six parameters $(\alpha i_1, \alpha i_2, \varphi i_1, \varphi i_2, \tau i_1, \tau i_2)$ with each $\alpha, \phi, \tau \in [0, 1]$. These determine, respectively, the amplitude phase and duty cycle of the walking signal sent to the first two legs every 30ms. The walking signal for the third segment is the inverse of that for the second so does not need independent control parameters. The hexapod controller is defined by the six parameters for each leg for a total of 36 parameters, and fully determines the hexapod gait. Using the hexapod simulator in [5] we define a CNMA forward function $hexapod(x) = y$ that takes a controller $x = [c_0, .., c_{35}]$ as input, simulates the hexapod gait for 5 seconds and outputs a vector $y = [speed_x, b_1, b_2, b_3, b_4, b_5, b_6]$ where $speed_x$ is the hexapod's $X$-axis displacement in meters divided by 5.0 and each $b_i$ is the fraction of the time leg $i$ was in contact with the ground.

The baseline controller in [5] for the hexapod is:

$$[c_0, .., c_{35}] = [1.0, 0.0, 0.5, 0.25, 0.25, 0.5, 1.0, 0.5, 0.5,\\ 0.25, 0.75, 0.5, 1.0, 0.0, 0.5, 0.25, 0.25, 0.5,\\ 1.0, 0.0, 0.5, 0.25, 0.75, 0.5, 1.0, 0.5, 0.5,\\ 0.25, 0.25, 0.5, 1.0, 0.0, 0.5, 0.25, 0.75, 0.5]$$

For this controller, $hexapod$ returns the output:

$$[speed_x, b_1, b_2, b_3, b_4, b_5, b_6] = [0.20\ 0.32\ 0.4\ 0.08\ 0.14\ 0.37\ 0.32]$$

The speed of 20cm/sec is a threshold for the hexapod in that speeds above this are hard to find with random search.

**Problem 1.** If a hexapod leg is broken then we would like to find a new controller that can achieve a speed of close to 20cm/sec while satisfying any new constraints on the movement. Let us assume leg 1 is broken. Then, we might constrain its contact with the ground to be the least of that of all the legs. The problem is now: maximize $speed$ subject to $hexapod([c_0, .., c_{35}]) = [speed, b_1, .., b_6] \wedge b_1 \leq b_2 \wedge b_1 \leq b_3 \wedge b_1 \leq b_4 \wedge b_1 \leq b_5 \wedge b_1 \leq b_6$ and stopping when the speed is close enough to 20cm/sec. As can be seen above, the baseline controller does not satisfy the constraint.

We can try searching over $[0,1]^{36}$ for a new controller. But, another option is to search just in the neighborhood of the baseline controller. We change the bounds for each field in the baseline controller to be within 0.1 of its current value, subject to the lower bound being at least 0 and the upper bound at most 1. In 280 function calls over 20 minutes, CNMA returns the following controller:

$$[c_0, .., c_{35}] = [0.95, 0.0, 0.6, 0.15, 0.35, 0.6, 0.9, 0.4, 0.45,\\ 0.15, 0.65, 0.55, 0.9, 0.1, 0.4, 0.15, 0.35, 0.4,\\ 0.9, 0.0, 0.6, 0.15, 0.65, 0.6, 0.9, 0.45, 0.4,\\ 0.35, 0.15, 0.6, 0.9, 0.0, 0.6, 0.15, 0.65, 0.4]$$

With associated output:

$$[speed_x, b_1, b_2, b_3, b_4, b_5, b_6] = [0.1998504, 0.25, 0.36, 0.32, 0.28, 0.32, 0.45]$$

As can be checked, this output satisfies the constraint. If we randomly search for a controller satisfying the constraint, even *in this reduced space around the baseline controller*, there were just 29 out of 100,000 samples satisfying the constraints with a speed greater than or equal to 0.199. This translates to an average of 3448 function evaluations, more than one order of magnitude (12.31 times) larger than 280.

**Problem 2.** We also explored how much we could improve upon the baseline controller if we could afford a much larger function evaluation budget. Again, we searched around the baseline controller. We just maximized $speed$ subject to $hexapod([c_0, .., c_{35}]) = [speed, b_1, .., b_6]$. Note we did not impose any constraints on the $b_i$. In 3670 function calls over 8 hours, CNMA improved the speed by 50% to produce a controller with a speed of 29.99cm/sec. The new controller is:

$$[c_0, .., c_{35}] = [1.0,0.0,0.6,0.35,0.3,0.4,1.0,0.4,0.4,\\ 0.35,0.65,0.6,0.9,0.1,0.4,0.35,0.35,0.6,\\ 1.0,0.1,0.6,0.35,0.85,0.4,1.0,0.4,0.4,\\ 0.15,0.3,0.4,1.0,0.0,0.6,0.35,0.65,0.45]$$

with associated output $[speed_x, b_1, b_2, b_3, b_4, b_5, b_6] = [0.299, 0.29, 0.32, 0.14, 0.2, 0.42, 0.27]$. There was not a single controller in 100,000 randomly generated samples in the reduced search space with a speed exceeding 29.99cm/sec.

## 8. Modeling nonlinear constraints and objective functions

We show how, despite using a MILP solver, CNMA allows $P$ and $\phi$ to be nonlinear. The idea is to move all nonlinear expressions into the CNMA forward function $F$, add output variables computing values of these, and express equivalent linear $P$ and $\phi$ in terms of these variables.

### 8.1 Modeling nonlinear constraints

To model a nonlinear constraint $P$, we add, for each nonlinear expression in $P$, an extra output to $F$ denoting the value of the expression and then express $P$ as a linear constraint on these outputs. For example, let $F(X) = Y$ where $X = [x_1, x_2]$, $Y = [v]$ and $v = x_1 + x_2$. Suppose $P(X, Y) = v^2 \geq 5$, a nonlinear constraint. CNMA's MILP solver cannot process this constraint. However, we can extend $F$ to compute an extra output $w = (x_1 + x_2)^2$ so that $Y = [v, w]$. Now, the nonlinear constraint is expressed as the linear $w \geq 5$ and solved with CNMA.

We use this idea to solve a nonlinear optimization benchmark problem [**14**]. This problem has 12 variables $X = (x_1, .., x_{11}, u), x_i \in [-1, 1], u \in [-1, 10]$. There are 10 constraints each of the form $0 \geq E_i(X)$ where each $E_i$ is a transcendental function of variables in $X$. The optimization problem is to minimize $u$ subject to the 10 constraints. An example of a constraint is:

```
0 >= (exp((x1-sin(0.0+1.0+1.0))*(x1-sin(0.0+1.0+1.0)))) +
0.5*(exp((x2-sin(0.0+2.0+2.0))*(x2-sin(0.0+2.0+2.0)))) +
0.3333333333333333*(exp((x3-sin(0.0+3.0+3.0))*(x3-sin(0.0+3.0+3.0)))) +
0.25*(exp((x4-sin(0.0+4.0+4.0))*(x4-sin(0.0+4.0+4.0)))) +
0.2*(exp((x5-sin(0.0+5.0+5.0))*(x5-sin(0.0+5.0+5.0)))) +
0.1666666666666666*(exp((x6-sin(0.0+6.0+6.0))*(x6-sin(0.0+6.0+6.0)))) +
0.14285714285714285*(exp((x7-sin(0.0+7.0+7.0))*(x7-sin(0.0+7.0+7.0)))) +
0.125*(exp((x8-sin(0.0+8.0+8.0))*(x8-sin(0.0+8.0+8.0)))) +
0.1111111111111111*(exp((x9-sin(0.0+9.0+9.0))*(x9-sin(0.0+9.0+9.0)))) +
0.1*(exp((x10-sin(0.0+10.0+10.0))*(x10-sin(0.0+10.0+10.0)))) +
0.09090909090909091*(exp((x11-sin(0.0+11.0+11.0))*(x11-sin(0.0+11.0+11.0)))) -
u;
```

To solve this problem in CNMA, we define $F(X) = Y$ where $X = [x_1, .., x_{11}, u]$ and $Y = [v_1, .., v_{10}]$, each $v_i = E_i(X)$. Define $P(X, Y) = (0 \geq v_1) \wedge .. \wedge (0 \geq v_{10})$. Finally, $\phi(X, Y) = u$. CNMA solved this problem to produce the solution:

$$X = (-0.025802716144530603, 0.267246588244859, 0.11409408476703223, 0.16516646437336022, -0.15582812349227032, -0.0434702545214761, 0.2699575598670672, 0.021578735032435736, 0.27952956951645413, 0.2537270238373449, 0.046348332349110455, 6.315250767638159)$$

$$Y = (-0.21884700655881772, -0.7617891125086622, -1.167333092823653, -0.10171508987575084, -1.2243184765747221, -0.9139871220406643, -0.04651122087569082, -0.46619849507939204, -1.3713955572721979, -0.16156764926343392)$$

$$\phi(X, Y) = 6.315250767638159$$

As can be seen, each number in $Y$ is less than or equal to 0 as the constraints require. The true minimum is 5.93. Since the mathematical expressions are visible in this problem, it is efficiently solved by derivative-based solvers such as LOQO at https://neos-server.org/neos/. This problem is used here to illustrate CNMA's capability to model nonlinear constraints and objective functions. Since it has a significant number of complex nonlinear constraints, it is also an interesting benchmark for derivative-free solvers.

## 8.2 Modeling nonlinear objective function

Similarly, if our objective were to minimize $\phi(X) = \sum_{n=0}^{11} X[i]^2$ i.e., the sum of squares of all input fields, we would make $F$ compute this sum as an extra output $sq$ in $Y$, define $\phi(X, Y) = sq$ and then minimize $\phi(X, Y)$ s.t. $F(X) = Y \wedge P(X, Y)$.

# 9. Designing an acrobot

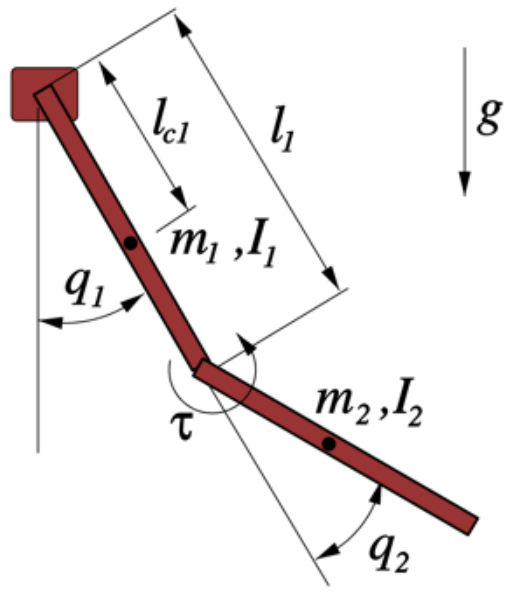

Figure 6. Acrobot schematic

We compare CNMA's performance on the two link Acrobot [16] system shown in Figure 6, with that of Nelder-Mead, Random Search, and Gaussian Processes. We find that CNMA needs only 125 function evaluations compared to 500+ with other solvers yet computes superior results.

The Acrobot is a two-link robot arm with a single actuator placed at the elbow. Initially, the links are hanging downwards. The Acrobot's goal is to execute a series of actions that vertically orients and balances both links. The Acrobot problem is well-studied, and is known to be challenging to solve. Several controllers exist to solve the Acrobot problem given a fixed system design. In this paper, we employ a continuous force variation of OpenAI Gym's implementation of the Acrobot problem, with maximum torque of $10\ \mathrm{N} \cdot \mathrm{m}$ and timestep of $dt = 0.2$ s; we employ iterative LQR as our controller. We keep the iLQR hyperparameters fixed. We do not solve or optimize for them as we did for the PID controller for the lunar lander.

In the above figure, system design variables $m_i$, $I_i$, and $l_{c_i}$ denote, respectively, the mass, moment of inertia, and center of mass location of link $i$. State variables $q_i$ denote the angular position of link $i$ while action variable $\tau$ represents the torque applied.

The CNMA function $F$ takes as input a system design vector $X = [m_1, m_2, I_1, I_2, l_{c_1}, l_{c_2}, l_1, l_2]$, runs iterative LQR on an Acrobot system with this vector and returns $t_{stabilize}$, the total time taken to balance the system. In the event that no solution is found, $F$ returns $\infty$. The state variables are internal to $F$. The design problem is to find a system design that minimizes $\phi(X, Y) = t_{stabilize}$ subject to $F(X) = Y \wedge P(X, Y)$ where:

$$
\begin{gathered}
P(X, Y) = \\
(I_1 = I_2 \wedge \\
0.1 \leq m_1, m_2 \leq 3.0 \wedge \\
0.1 \leq I_1, I_2 \leq 3.0 \wedge \\
0.1 \leq l_1, l_2 \leq 3.0 \wedge \\
0.3 \leq l_{c_1}, l_{c_2} \leq 0.7)
\end{gathered}
$$

The constraint $I_1 = I_2$ reflects OpenAI/Gym's implementation of the Acrobot problem. CNMA finds the following solution:

$$
\begin{gathered}
[m_1, \quad m_2, \quad I_1, \quad I_2, \quad l_{c_1}, \quad l_{c_2}, \quad l_1, \quad l_2, \quad t_{stabilize}] \\
= \\
[3, \quad 0.1, \quad 0.1, \quad 0.1, \quad 0.3, \quad 0.7, \quad 0.1, \quad 0.1, \quad 2.8]
\end{gathered}
$$

## 10. Designing a cartpole

This problem is also inspired by OpenAI/Gym's cartpole [**16**]. The black cart is fixed to a grey horizontal line and the brown pole is attached to the cart at a joint. When a horizontal force is applied to the cart, the cart moves left or right causing the pole to rotate about the joint. A cartpole controller applies a force at points in time to bring the pole into the upright or balanced position i.e., pointing upwards. The design problem we solve is finding the parameters of a cartpole that minimizes the time an iLQR controller takes to bring the pole into the upright position subject to the constraint that the cart never moves more than a fixed distance away from the center.

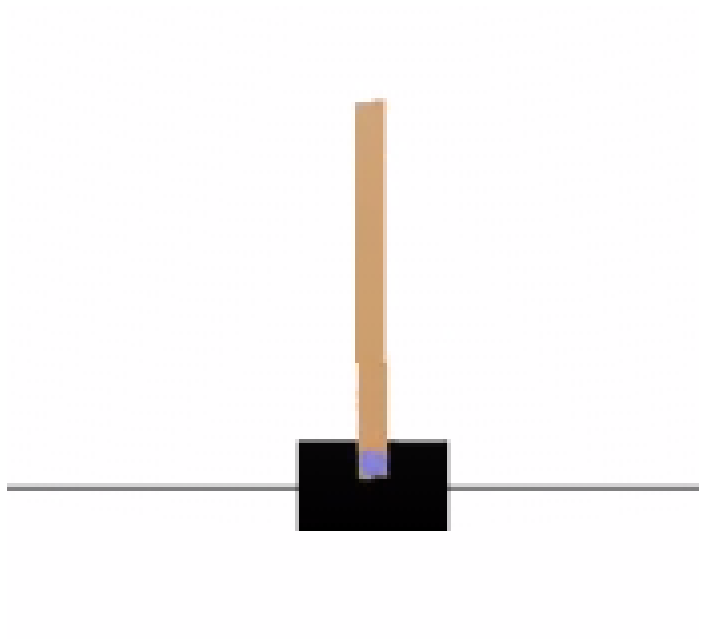

Figure 7. A Cartpole

The CNMA function $F$ takes as input $X = [m_c, m_p, l_p, max_f]$ where the fields are, respectively, the mass of the cart, mass of the pole, length of the pole and the magnitude of the force that is applied to the cart at each step in the sequence. $F$ then simulates an iLQR controller attempting to drive the pole into the upright position. It outputs $Y = [t_{upright}, max_{disp}]$ where $t_{upright}$ represents the time taken to balance the cartpole while $max_{disp}$ represents the maximum distance of the cart from the center at any point during the simulation. If the controller is unsuccessful, $t_{upright} = \infty$. The objective function to minimize is $\phi(X, Y) = t_{upright}$ and the the constraint, including bounds, is:

$$P(X,Y) = \begin{Bmatrix} 0.1 \le m_c \le 5 \\ 0.1 \le m_p \le 0.5 \\ 0.8 \le l_p \le 3 \\ 1 \le max_f \le 10 \\ max_{disp} \le 2.5 \end{Bmatrix}$$

CNMA produces the following solution:

$$[m_c, \quad m_p, \quad l_p, \quad max_f, \quad t_{upright}, \quad max_{disp}] = [0.9721940748558899, 0.1, 1.2181869634032596, 8.779277145488992, 7.199999999999999, 0.6735909024745416]$$

## 11. Designing a parallel parking maneuver

We show how to use CNMA to design a maneuver inspired by that of using Gompertz curves for parallel parking, from Toyota [1]. A general principle is to model a maneuver as a parametric plan and then, ask: if the maneuver were performed with that configuration for an appropriate amount of time, would safety be preserved? Encode simulator as $F(x) = y$ where $x$ models the plan parameters and $y = true/false$. Now use CNMA to find $x, y$ such that $F(x) = y \wedge y = true$.

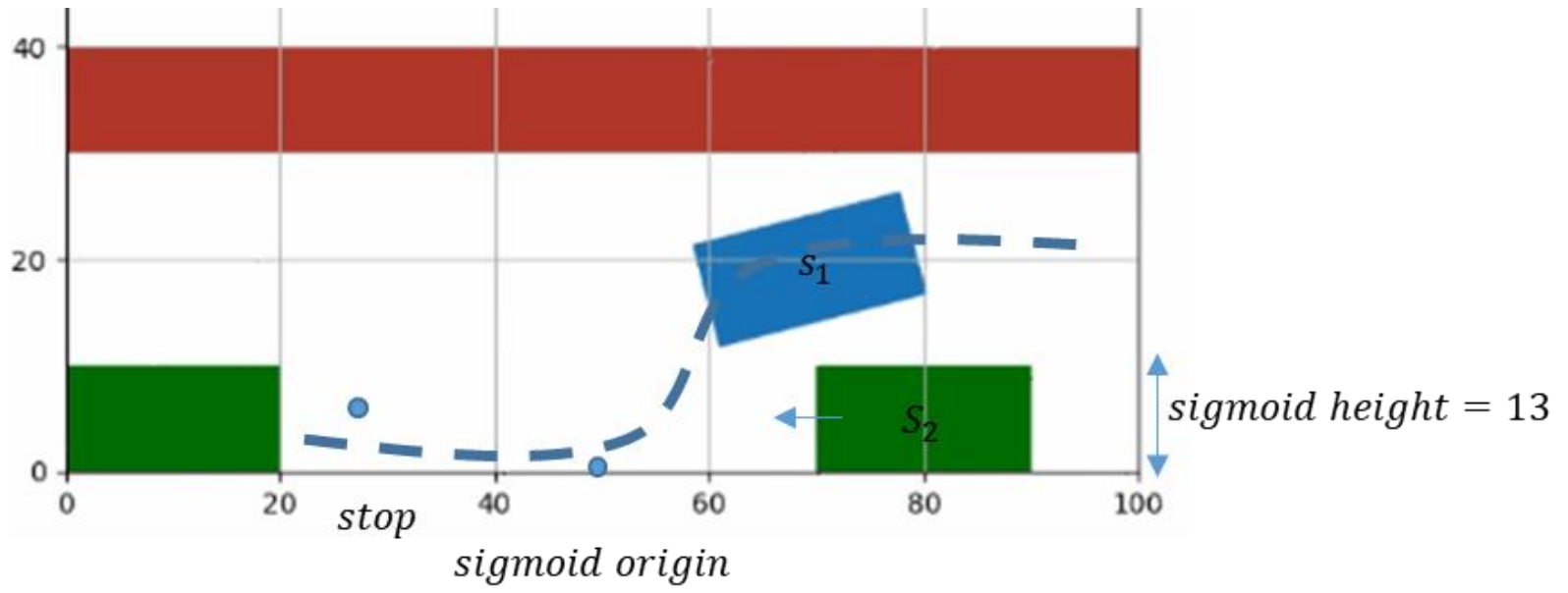

Figure 8. Finding Parameters of Sigmoid Curve For Safe Parallel Parking With Moving Obstacles

In Figure 8, we encode the function $safe(s_1, s_2, damp, stop)$ that returns $true/false$ depending on whether the following conditions are satisfied by the blue car over a long enough period of time:

- Starts at (100, 13)
- Moves at speed $s_1$ along a sigmoid $y(x) = \frac{13}{1+e^{-damp*x}}$ with origin $(50,\ 0)$
- Stops at $(stop,\ 5)$ where the derivative of the curve is close to 0, so car is straightened
- Never collides with green car at right starting at $(100,\ 5)$ moving at speed $s_2$
- Never collides with road divider at top
- Never collides with car at left

Variable bounds are: $s_1 \in [0, 5]$, $s_2 \in [0, 5]$, $damp \in [0, 1]$, $stop \in [0, 100]$. We now use CNMA to find $s_1, s_2, damp, stop$ such that $safe(s_1, s_2, damp, stop)$ subject to additional constraints: $s_2 = 0.1, stop \geq 30$. CNMA only needed to call the $safe$ function 104 times and solved the constraint in a few seconds. The solution was $[s_1 = 4.27, s_2 = 0.1,\ damp = 0.22, stop = 30.0]$. Since $s_2$ is moving and $stop \geq 30$, the blue car has to travel at a rather high speed and bend rather sharply to get ahead of the the car at right and stop without collision. The neural network architecture used was $(4, 20, 20, 1)$ with (20, 20) being the hidden layers. No comparison with other solvers was made for this as yet.

## 12. Experimental results

CNMA's neural networks always used a single hidden layer. The number of function evaluations (calls) for all solvers is the sum of those made to generate initial samples and those made at all of the iterations. A sample is a solution if it satisfies all constraints in the optimization problem. Python packages used for neural networks, Gaussian processes and Nelder Mead were, respectively, scikit-learn, skopt and scipyopt. A commercial MILP solver was used. The acquisition function for Gaussian processes was Expected Improvement.

### 12.1 Designing wave energy-propelled marine robot

**CNMA**

- Found solution with boat speed of 3.76m/s.
- Number of function calls = 249.
- Number of initial samples = 30.
- Neurons in hidden layer = 10.
- Time taken ≈ 5 minutes

**Random search**

- Only 1 out of 10,000 samples satisfied constraints with boat speed of 2.57 m/s. Time taken ≈12 minutes with parallel processing.

**Gaussian processes**

- Could not find a solution after 50 function calls. Solver terminated after this because

it began to generate the same set of new samples at each subsequent iteration. The forward function failed to terminate when evaluating these samples.

**Nelder Mead**

- Did not find a solution after 249 calls. Solver terminated on its own because it converged to within its tolerance, but the optima did not satisfy the constraints.

## 12.2 Robot system-controller codesign

**CNMA**

- Found solution with reward 447.78.
- Number of function calls = 246.
- Number of initial samples = 100.
- Neurons in hidden layer = 30.
- Time taken to find best solution = 5 minutes.

**Random search**

- Best solution out of 10,000 samples had a reward of 423.41.
- Time taken ≈ 2 minutes with parallelism.

**Gaussian processes**

- Best solution had a reward of 365.
- Solver terminated after 125 function calls because it began to generate the same set of new samples at each subsequent iteration.
- Time taken ≈ 2 minutes.

**Nelder-Mead**

- Best solution had a reward of 411.98.
- Number of function calls = 757. The best solution was found after only 246 function calls.
- Time taken ≈ 1 minute.

## 12.3 Designing hexapod gait

**Problem 1**

**CNMA**

- Found solution with a speed of 19.99cm/sec.
- Number of function calls = 280.
- Number of initial samples = 2.
- Neurons in hidden layers = 30.
- Time taken = 20 minutes.

**Random search**

- Just 29 out of 100,000 samples satisfied the constraints with a speed greater than or equal to 19.99cm/sec. This translates to an average of 3448 function evaluations to find a satisfying solution with a speed greater than or equal to 19.99. This is more than one order of magnitude (12.31 times) larger than 280 with CNMA.
- Time taken = 20 minutes.

**Gaussian Processes**

- Found solution with speed of 18.33 cm/sec.
- Number of function calls = 522.
- Time taken = 10 hours (was manually terminated after this)

**Nelder Mead**

- Was not able to find a solution that satisfies all the constraints.
- Number of function calls = 2737
- Time taken = 20 minutes (was manually terminated after this)

**Problem 2**

**CNMA**

- Found solution with speed of 29.99 cm/sec
- Number of function calls = 3670
- Number of initial samples = 2
- Neurons in hidden layer = 30
- Time taken = 8 hours

**Random**

- There was not a single randomly generated controller out of 100,000 samples with a speed exceeding 29.99cm/sec

**Gaussian Processes**

- Found solution with speed of 24.26 cm/sec
- Number of function calls = 843
- Time taken = 18 hours (was manually terminated after this)

**Nelder Mead**

- Found solution with speed of 20.63 cm/sec
- Number of function calls = 3968
- Time taken = 30 minutes. (was manually terminated after this)

## 12.4 Modeling nonlinear constraints and objective functions

**CNMA**

- Best solution is $u = 6.315$
- Number of function calls = 104
- Time taken = 80 seconds
- Number of neurons in hidden layer = 35.
- Number of initial samples = 2.

**Random search**

- Best solution is $u = 7.051$
- Number of function calls = 100000
- Time taken = 173 seconds

**Gaussian processes** (terminated after 73 iterations)

- Best solution is $u = 276.135$
- Number of function calls = 1460
- Time taken = 11 hours.

**Nelder-Mead**

- Best solution is $u = 7.11581$.
- Number of function calls = 315
- Time taken = 3.815 seconds

## 12.5 Designing an acrobot

**CNMA**

- Best solution was $t_{stabilize} = 2.8$ sec.
- Number of function calls = 150.

- Time taken = 41.3 min.
- Number of initial samples = 20.
- Number of neurons in hidden layers = 10.

**Random search**

- Best solution was $t_{stabilize} = 5\ sec$
- Number of function calls = 1000.
- Time taken = 100.1 minutes

**Gaussian processes**

- Best solution was $t_{stabilize} = 3.2$ sec.
- Number of function calls = 600.
- Time taken = 295.6 min.

**Nelder-Mead**

- Best solution was $t_{stabilize} = 4.6$ sec.
- Number of function calls = 1163.
- Time taken = 118.9 min.

## 12.6 Designing a cartpole

**CNMA**

- Best solution was $t_{upright} = 7.2$ seconds.
- Number of function calls = 289.
- Time taken = 2000 seconds.
- Number of initial samples = 100.
- Neurons in hidden layer = 30.

**Random search**

- Best solution was $t_{upright} = 7.95$ seconds.
- Number of function calls = 5000.
- Time taken = 10,000 seconds

**Gaussian processes**

- Solver manually terminated because it got into a loop, generating the same sample.
- Best solution was $t_{upright} = 8.85$ seconds.
- Number of function calls = 182.
- Time taken = 3600 seconds.

**Nelder-Mead**

- Solver terminated itself because it converged to within its tolerance.
- Best solution was $t_{upright} = 8.55$ seconds.
- Number of function calls = 200.
- Time taken = 600 seconds.

# 13. Conclusions and discussion

Central to designing many robots and their controllers is the need to solve a constrained blackbox optimization problem. This paper has presented CNMA, a new method of solving this problem and evaluated it to designing several diverse robotics systems and their controllers: wave-energy propelled boat, system-controller codesign for the OpenAI/Gym lunar lander, hexapod gait repair, cartpole, acrobot and a parallel parking maneuver. In addition, a benchmark nonlinear optimization problem was solved. These problems are of between 6 and 36 real-valued dimensions with complex constraints spanning both input and output variables. The paper shows that CNMA surpasses several derivative-free optimization tools: Nelder Mead, Gaussian Processes and Random Search with respect to the metric of number of function evaluations. This metric is a critical one when blackbox evaluation is expensive.

CNMA also surpasses optimization tools in expressive power. It allows complex constraints, including recursive ones that govern both input and output, as constraints in a MILP language. It eliminates the need to encode constraints as hard-to-devise penalty or barrier functions. Furthermore, the MILP solver exploits the special nature of constraints to efficiently solve these. Traditional methods do not make any use of the special nature of constraints, treating constraint satisfaction as another optimization problem and thus suffer a loss in efficiency as the evidence above indicates.

Like other derivative-free methods CNMA uses a surrogate function to compute the next sample to evaluate in search of a solution. The difference is that the surrogate function is a neural network and its equivalent MILP is used to suggest the next relevant point to sample in its learning-from-failure step. Since MILP solvers are optimized to constraint solving they are a natural candidate for evaluating relevance to the *constraint* part of the problem. Thereby, the number of samples needed to solve problems is many orders of magnitude less than that needed to learn the functions over their entire domain. An extreme case was the hexapod where CNMA started with just two initial samples to compute a speed that was hard or impossible to surpass by traditional solvers. Interestingly, in CNMA it is the constraint solver that drives optimization rather than optimization driving constraint satisfaction (as with penalty/barrier methods).

While some examples in this paper address controller design, CNMA is applicable broadly to generic nonlinear optimization problems such as designing the structure of the robot or co-design of structure, material, process, and controllers. Note that in the design of the wave-energy-propelled robot, no controller was designed.

If a large and deep neural network is needed to model a complex robotics function, its MILP equivalent may not be efficiently solvable. However, CNMA does not need to model the function in its entire domain. It only needs to model it in the part of the domain relevant to solving the constrained optimization problem. If this region is not too complex, a smaller neural network is adequate so its MILP equivalent could be efficiently solvable. This region is automatically computed by CNMA. As we have seen, even for the 36 dimension problem, single hidden layer neural networks with 30 neurons was adequate. In fact, a large or deep neural network may be detrimental to performance as it would overfit to the small number of points at which CNMA samples. Note that Nelder-Mead only makes linear approximations of the functions in the regions it searches in. Thus, it is not a requirement for blackbox optimization that the surrogate function model the function over its entire domain.

**Acknowledgements.** We appreciate the assistance of Professor Antoine Cully in providing us with the hexapod simulator and the insights into his methods. We also appreciate feedback from Professor Robert Vanderbei, Professor John Chinneck, Professor Nick Sahinidis, Dr. Matthias Poloczek, Professor Jaime Fisac, Mr. Nicholas Kraus and Mr. Taylor Njaka.